\documentclass[letterpaper, 10 pt, journal, twoside]{IEEEtran}

\usepackage{amsmath}
\usepackage{amssymb}
\usepackage{mathtools}
\usepackage{stackrel}
\usepackage{graphicx}
\usepackage{threeparttable}
\usepackage{xcolor}
\usepackage{subcaption}
\usepackage{algorithm}
\usepackage{algpseudocode}
\usepackage{multirow}
\usepackage[utf8]{inputenc}
\begin{document}
%

\title{TacGNN: Learning Tactile-based In-hand Manipulation with a Blind Robot using Hierarchical Graph Neural Network}

\author{Linhan~Yang$^{1}$, Bidan~Huang$^{2 \dag}$, Qingbiao~Li$^{3}$, Ya-Yen~Tsai$^{4}$, Wang~Wei~Lee$^{2}$, Chaoyang~Song$^{5}$, Jia~Pan$^{6}$%
\thanks{Manuscript received: January 20, 2023; Accepted March 17, 2023.}%
\thanks{This paper was recommended for publication by Editor Ashis Banerjee upon evaluation of the Associate Editor and Reviewers' comments.}%

\thanks{$\dag$ denotes the corresponding author.}%
\thanks{$^{1}$Linhan Yang is with the Department of Mechanical and Energy Engineering, Southern University of Science and Technology and the Department of Computer Science, University of Hong Kong, Hong Kong. This study was conducted during his internship at Tencent Robotics X. 
        {\tt\footnotesize  yanglh@connect.hku.hk}}%
\thanks{$^{2}$B. Huang, Wang Wei Lee are with Tencent Robotics X, China {\tt\small bidanhuang@tencent.com, wwlee@tencent.com}}%
\thanks{$^{3}$Qingbiao Li is with the Department of Computer Science and Technology, University of Cambridge, Cambridge, United Kingdom
        {\tt\footnotesize  ql295@cam.ac.uk}}%
\thanks{$^{4}$Y.-Y. Tsai is with the Hamlyn Centre for Robotic Surgery, Department of Computing, Imperial College London, SW7 2AZ, London, UK {\tt\footnotesize y.tsai17@imperial.ac.uk}}%
\thanks{$^{5}$Chaoyang Song is with the Department of Mechanical and Energy Engineering, Southern University of Science and Technology, Shenzhen, Guangdong
518055, China.
        {\tt\footnotesize  songcy@ieee.org}}%
\thanks{$^{6}$Jia Pan is with the Department of Computer Science, University of Hong Kong,Hong Kong 999077.
        {\tt\footnotesize  jpan@cs.hku.hk}}%
\thanks{Digital Object Identifier (DOI): see top of this page.}%
}

\markboth{IEEE Robotics and Automation Letters. Preprint Version. Accepted March, 2023}
{Yang \MakeLowercase{\textit{et al.}}: Blind Robot Manipulation} 

\maketitle

\begin{abstract}
In this paper, we propose a novel framework for tactile-based dexterous manipulation learning with a blind anthropomorphic robotic hand, i.e. without visual sensing. First, object-related states were extracted from the raw tactile signals by a graph-based perception model - TacGNN. The resulting tactile features were then utilized in the policy learning of an in-hand manipulation task in the second stage. This method was examined by a Baoding ball task - simultaneously manipulating two spheres around each other by 180 degrees in hand.
We conducted experiments on object states prediction and in-hand manipulation using a reinforcement learning algorithm (PPO). Results show that TacGNN is effective in predicting object-related states during manipulation by decreasing the RMSE of prediction to 0.096cm comparing to other methods, such as MLP, CNN, and GCN. Finally, the robot hand could finish an in-hand manipulation task solely relying on the robotic own perception - tactile sensing and proprioception. In addition, our methods are tested on three tasks with different difficulty levels and transferred to the real robot without further training. https://sites.google.com/view/tacgnn
\end{abstract}


\IEEEpeerreviewmaketitle

\section{Introduction}
\label{sec:intro}

    \IEEEPARstart{E}{nabling} human-level manipulation in robots has always been a roboticist's dream~\cite {bicchi2000hands}. The human hand is unparalleled in its ability to perform dexterous object manipulation. While the outstanding capabilities of the human hand may be largely attributed to the many degrees of freedom afforded by its highly evolved anatomy, the availability of high definition tactile feedback is equally vital. Tasks such as grasping, in-hand manipulation, and finger gaiting cannot be efficiently achieved without tactile feedback, since the in-hand object state is derived from contact area distribution while the breaking and re-establishing of contacts signal key transitional events in the process of manipulation~\cite{Johansson2009}. Understandably, achieving human-like object manipulation in robotic hands is a grand challenge.
    
    Unlike humans, robotic manipulation has long relied on vision to interpret the interactions between the object and the hand. Robotic vision systems typically involve precise calibration, multiple cameras, complicated algorithms, or expensive motion capture systems~\cite{andrychowicz2020learning,nagabandi2020deep,akkaya2019solving}. Unfortunately, vision is not a direct observation of force and thus cannot provide the contact information needed for dexterous manipulation. Therefore, though vision-based methods~\cite{andrychowicz2020learning} have made commendable progress in recent years, manipulation remains an open challenge as the use of tactile information has not yet been fully explored.
    
        \begin{figure}[t]
          \centering
          \includegraphics[width=0.5\textwidth]{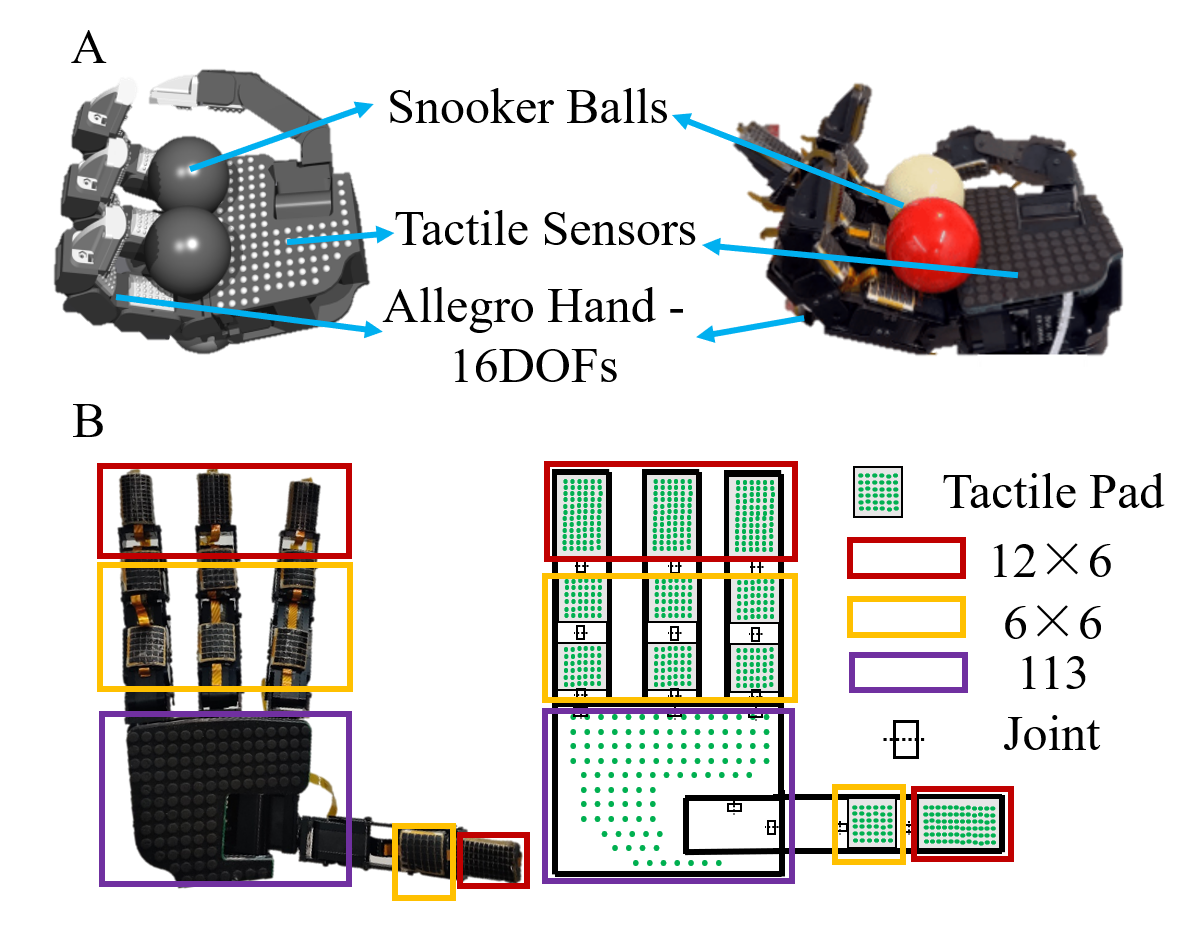}
          \caption{Hardware setup. (A) An Allegro Hand covered with distributed tactile sensors manipulates two in-hand snooker balls in simulation (left) and real-world (right). (B) Sensor map on Allegro Hand.
    }
          \label{fig:overview}
          \vspace{-2mm}
       \end{figure}
       
    One major obstacle in the integration of tactile signals for manipulation is the difficulty in developing 
    an analytical model that captures the relationship between time-variant tactile sensor readings and hand motion. Recently, learning-based approaches have been explored to directly learn from experience or demonstrations to achieve dexterous manipulation~\cite{akkaya2019solving, sundaram2019learning}. Unlike other sensing modalities such as force/torque sensing, where the signals can simply be modeled as a time-series vector~\cite{lee2019making}, large area electronic skins typically generate tactile signals of high dimensionality which are often sparse and not well structured~\cite{Lee2015}. The sensors may also vary in shape and size depending on the hardware used~(Fig.~\ref{fig:overview}). The structure of the neural network is thus critical to the efficacy of learning-based manipulation using tactile feedback.
    
    Earlier works have attempted to model high-resolution tactile inputs using Convolutional Neural Network (CNN) ~\cite{sundaram2019learning,funabashi2020stable}. However, CNN accepts only rectangular inputs of predefined sizes,  necessitating the resizing of tactile inputs which discards the inherent spatial relationship between taxels.
    To overcome this limitation, Graph Convolutional Network (GCN)~\cite{kipf2016semi} has been introduced to encode and interpret tactile information~\cite{hu2021living, funabashi2022multi}. Although GCN could handle unstructured tactile signals, the adjacency matrix of the graph, such as the number of nodes and connections between nodes must be specified from the start and remain fixed thereafter. As a result, the network will have difficulty encoding the time-variant spatial relationship of tactile signals that arise from the diverse poses the robot hand may adopt during manipulation. The GCN will also not work for systems where the tactile sensors are configured differently, thus restricting the applicability of the solution across platforms.
    
    Unlike the aforementioned approaches, we treat tactile signals as a point set, where each point represents the 3D position of an activated tactile sensor. This tactile point set comes with non-uniform density in different areas as the contact areas are unevenly distributed on the hand. The tactile point set is interpreted by a hierarchical Graph Neural Network to capture features at different levels~\cite{qi2017pointnet++}. This tactile graph, referred to as TacGNN, is dynamic, i.e. the connectivity and node number of the graph changes as the fingers are moved and the contact areas vary. Unique to TacGNN is its ability to accommodate any sensor configuration and efficiently capture the spatial relationship between sensors. With the object pose predicted by TacGNN, we trained an autonomous control policy for an in-hand manipulation task using an on-policy Reinforcement Learning (PPO) algorithm only relying on the robotic perception (Section~\ref{sec:method}).
    
    \textbf{Contributions.} Overall, we propose a framework for learning and utilizing tactile
    features for in-hand manipulation tasks. Our main contributions are summarized as follows:
    \begin{enumerate}
        \item We proposed a novel method-TacGNN to capture tactile features from distributed tactile sensing, which outperforms prior methods like MLP, CNN, GCN on the object pose prediction experiments.
        
        \item We achieved a complex manipulation task - a Baoding ball task solely relying on the predictive output from TacGNN with an on-policy reinforcement learning algorithm.
        
        \item We tested our method on easy, moderate, and difficult tasks achieving 94.71\%, 88.57\%, and 79.78\% success rates, which greatly surpasses prior methods. 
        
        \item We transferred the policy learned in simulation onto real robots, demonstrating real-world results that are consistent with the simulation. 
    \end{enumerate}

\section{Related Work}
\label{sec:related}

    \subsection{Learning-based approach for in-hand manipulation}
    As the mechanism design of the hand becomes increasingly dedicated and complex, it is getting harder to design an analytical model~\cite{bhatt2022surprisingly} to control the hands for various tasks. The rise of deep learning provides researchers with huge potential to achieve dexterous manipulation by experiences like humans~\cite{akkaya2019solving, charlesworth2021solving, niu2023goats}. 
    Existing work has been using the benefit of visual data from a camera or motion capture system for dexterous manipulation~\cite{nagabandi2020deep, hu2021living, bhatt2022surprisingly}. For example, researchers from OpenAI propose model-free RL using vision data from a camera to solve Rubik's cube~\cite{akkaya2019solving}. Bhatt et al.~\cite{bhatt2022surprisingly} proposed a simple model without RL with surprising robustness performance for manipulation tasks for various shapes, sizes, and weights of objects based on a motion capture system. However, humans can achieve dexterous manipulation purely based on a tactile feeling from hands without any visual input. Such extra constraints in dexterous manipulation yield active research in processing tactile data and control algorithms for manipulation based on these data.

\subsection{Tactile-based approaches}
    Researchers have been investigating vector-based and image-based approaches to process tactile data~\cite{sundaram2019learning, funabashi2018object, yang2021learning, chen2023polymer}. 
    Lee et al.~\cite{lee2019making} simply model tactile signals as a vector. This approach is particularly useful for robot joint
    proprioceptive force/torque.
    Funabashi et al.~\cite{funabashi2018object} proposed a simple feed-forward CNN by considering the tactile sensing data as an image. However, the tactile sensor only obtained data when contacting the object, resulting in spatially sparse sensor reading as the object moves. Sundaram et al.~\cite{sundaram2019learning} propose a similar CNN-based method using distributed tactile sensors for object classification. They first reshape the tactile signals to a rectangular size to accommodate CNN input, which destroys the inherent spatial relationship. This brought difficulty to processing tactile sensor data based on CNNs approaches. 
    
    \subsection{GNN for in-hand manipulation}
    Recent research from molecular biology, to multi-robot path planning~\cite{wu2020comprehensive} have demonstrated promising performance of GNNs to capture the interaction between a node in the graph for node and edge classification and message passing. This property yields the rising research in capturing tactile sensor data using GNN~\cite{funabashi2022multi, garcia2019tactilegcn} for in-hand manipulation. The properties of GNNs in capturing inter-node reactions inspired researchers to use GNN as a control algorithm. 
    
    The strength of GNNs has been recently utilized in tactile sensing.  
    Garcia-Garcia et al.~\cite{garcia2019tactilegcn} proposed a GCN-based approach to capture taxel reading from BioTac SP taxel to binary classify grasps as stable or slippery ones. 
    Funabashi et al.~\cite{funabashi2022multi} proposed a GCN-based pipeline to extract geodesical features from the tactile data from distributed taxels.
    However, their work does not consider how the topology of the graph changes as the taxel reading changes.  
    Since the matrix dimensions of both input and output are fixed, GCN can only process the graph as a whole, and it is hard to process the subset of the whole graph.
    
        \begin{figure*}[!tp]
          \centering
          \includegraphics[width=0.95\textwidth]{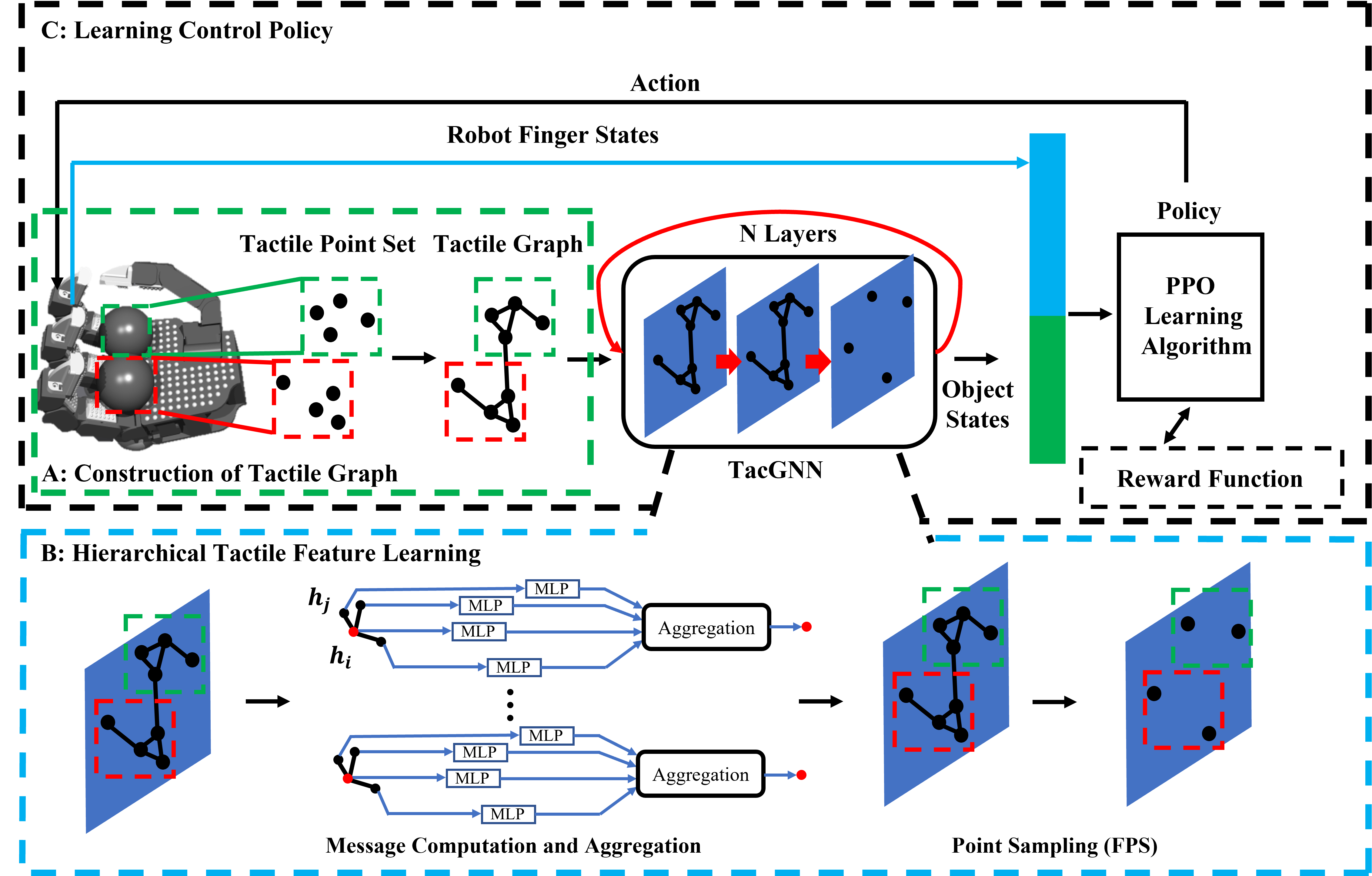}
          \caption{Schematic of the proposed control method for a Baoding ball task - simultaneously manipulating two spheres around each other by 180 degrees in hand. (A) First, tactile information is treated as a point set (3D position). Then, a tactile graph is generated based on the position of each tactile sensor. (B) Learning tactile features. First, we compute the embedding of each graph node from the neighborhood and aggregate them. Then, we sample points via farthest point sampling (FPS). These two steps are repeated in several layers.  (C) Finally, the object states predicted by TacGNN are concatenated to other robot states and fed into the control policy (PPO). The control policy generates an action based on the state vector and an update based on the reward function.
    }
          \label{fig:method}
          \vspace{-3mm}
       \end{figure*}
   
\section{Methodology}
\label{sec:method}

    First, we will introduce the basic concept of GNN in Section~\ref{sec:GNN}, and then Section~\ref{sec:TacGNN} explains how to use TacGNN to interpret object states from distributed tactile sensing. This model is called the perception model. All other methods, such as MLP, CNN, GCN are used as the perception model for the ablation study in Section~\ref{sec:experiment}.
    Then, the reinforcement learning formulation is demonstrated in Section~\ref{sec:rl}. With the learning algorithm, we learn a continuous control policy for an in-hand manipulation task. Finally, in Section~\ref{sec:workflow}, we summarize the perception model and control policy and show the workflow of the training process.

\subsection{Graph Neural Network}
\label{sec:GNN}
    In this section, we briefly review the concepts of graph neural networks. A graph typically consists of nodes and edges. A GNN is an optimizable transformation on all attributes of the graph (nodes, edges) that preserves graph symmetries (permutation invariances).~\cite{sanchez-lengeling2021a}
    Inspired by the documentation of PyTorch Geometric~\cite{Fey/Lenssen/2019}, we can generalize the convolution operator on images into the graph convolution operator that can aggregate information from the neighborhood or operate on a message-passing scheme. By defining node features of node $i$ in layer $(l-1)$ as $\mathbf{h}^{l-1}_{i}\in R$ and edge features from node $j$ to node $i$ as $\mathbf{e}_{j,i}\in R^{D}$, message passing graph neural networks can be described as
    \begin{equation}
    \mathbf{h}^{(l)}_{i} = \gamma^{(l)}\left(\mathbf{h}^{(l-1)}_{i}, \mathcal{A}_{j\in \mathcal{N}_{\text{in}}(i)} \phi^{(l)}( \mathbf{h}^{(l-1)}_{i}, \mathbf{h}^{(l-1)}_{j}, \mathbf{e}_{j, i})\right) 
    \end{equation}
    where $\mathcal{A}$ is a general symbol for any aggregation function that holds the property of differentiable, permutation invariant, for example, sum, mean, or max. Besides, we use $\gamma$ and $\phi$ to define differentiable functions such as MLPs (Multi-Layer Perceptron).

\subsection{Tactile Graph Neural Network (TacGNN)}
\label{sec:TacGNN}

    In this section, we introduce our proposed perception model - Tactile Graph Neural Network(TacGNN): first, we construct a tactile graph based on the position of each tactile. Second, we adopt a 3-layer hierarchical graph neural network to interpret tactile features.
    
\subsubsection{Graph Construction}
    As shown in Fig.~\ref{fig:method}, we define tactile information as a point set, $P = \{p_1, p_2, p_3, ..., p_{n}\} $, where $p_i = (x_i)$ is a taxel with 3D coordinates $x_i\in R^3$. 
    
    Given the point cloud P, we construct a tactile graph $G = (P, E)$ by finding the K nearest neighbors for each tactile node which is called the kNN search algorithm. $E$ denotes the edges connecting neighbors. k is 3 in this paper.

    A prominent feature of our method is that only activated tactile sensors are considered. As a result, the number of graph nodes varies depending on the interactive state between the robot hand and objects. Therefore, our graph is reconstructed in every single step. This approach can be implemented thanks to the fact that our method only propagates information between local nodes and does not need to consider the structure of the entire graph. The details about the tactile learning model are shown in the next subsection.
    
\subsubsection{Hierarchical Tactile Feature Learning}

    Naturally, the activated tactile sensors are sparsely distributed, which means our tactile point set comes with non-uniform density in different areas. Only the sensors in the contact area are activated. In our project, one contact between the ball and hand normally results in 4-5 activated tactile sensors due to compliance. The compliance in simulation is based on the real robot. 

    Such non-uniformity introduces a significant challenge for point-set feature learning. Inspired by CNN which is able to progressively capture features at increasingly larger scales along a multi-resolution hierarchy using the Pooling operation. At lower levels, neurons have smaller receptive fields whereas at higher levels they have larger receptive fields. Therefore, we adopt a hierarchical graph neural network with a similar function to process this tactile distribution~\cite{qi2017pointnet++},\cite{shi2020point}.

    Here is the basic idea of our model: we first compute the message propagation within a small neighborhood to extract features in local connection; then, the graph is downsampled into a sparse graph so that we can extract the feature on a larger scale. This process is repeated until we obtain the features of the whole tactile graph. An object state prediction MLP is used to get the final output.
    Our hierarchical structure is composed by a set abstraction operation - Point sampling as shown in Fig.~\ref{fig:method}(B). 
    
    In summary, our whole hierarchical network consists of three GNN layers, and each layer is made of three stages: Message Computation, Message Aggregation, and Point Sampling. An MLP is added in the end to get the object state. 

    \noindent{\textbf{Message Computation}}.
    In this stage, we compute the passed message for each edge using an MLP with hidden features and relative positions as input. The input dimension is $C\times2+3$, and the output dimension is $C$ where $C$ denotes the feature channels of each node. We further discuss the choice of $C$ in Section~\ref{sec:manipulation}.
    \begin{equation}
        \mathbf{h}_{i,j}^l = \phi^{(l)}\left(\mathbf{h}_i^{l}, \mathbf{h}_j^{l}, x_j-x_i\right) 
    \end{equation}
    where $h_j^{l}$ is the hidden feature of node j at propagation layer $l$, $x_j, x_i$ is euclidean coordinates of point $i,j$ and $h_{i,j}^l$ is the message propagated by $j$ to $i$ ($j$ is one of the neighbours of $i$). The coordinates of points are first translated into a local frame relative to the centroid point: $x_j-x_i$. Using relative coordinates is more efficient in capturing point-to-point relations in the local region~\cite{qi2017pointnet++}.
    
    \noindent{\textbf{Message Aggregation}}.
    For each node, we get $n \times C$ features from neighbours. However, the number of each node is different. To maintain dimensional consistency, we aggregate the passed message for each point. The input dimension is $n \times C$, and the output dimension is $C$ where $C$ denotes the feature channels of each node.
    \begin{equation}
        \mathbf{h}_{i}^{l+1} = \mathcal{A}\left({\mathbf{h}_{i,j}^l \mid j\in \mathcal{N}_{\text{in}}(i)}\right)
    \end{equation}
    where $h_{i,j}^l$ is the message propagated by $j$ to $i$ ($j$ is one of the neighbours of $i$) and $\mathcal{N}_{\text{in}}(i)$ includes all the neighbours of node $i$, and
    $h_{i}^{l+1}$ denotes the hidden feature of node $i$ at propagation step $l+1$. The aggregation function $\mathcal{A}$ used is max in this paper
    
    \noindent{\textbf{Point Sampling}}.
    To extract a hierarchical feature, we downsample the point set via farthest point sampling (FPS)~\cite{moenning2003fast}. Given input points $\{p_1, p_2, ..., p_n\}$, we use FPS to choose a subset of points $\{p_{i1}, p_{i2}, ..., p_{im}\}$ such that $p_{ij}$ is the most distant point from the set  $\{p_{i1},p_{i2},...,p_{ij-1}\}$ with regard to the rest points. Using an iterative FPS allows the network to extract from local features to global features. The input is a graph with $N\times C$ dimension and the output is a graph with $N' \times C$ dimension where $N' < N$. In this paper, $N' = 0.5\times N$
    
    \noindent{\textbf{Object State Prediction}}.
    Finally, we aggregate the whole graph nodes to get a graph-level feature. The input is the final graph with $N\times C$ dimension and the output is one node with $C'$ channels feature. The aggregation function $\mathcal{A}$ used is max in this paper. Then, an MLP is used to predict the object states, i.e. position and orientation of objects. The 3D position of two balls - a 6D vector- is predicted in the manipulation task. The training details will be shown in Section~\ref{sec:workflow}.
    \begin{equation} 
        \hat{o} = MLP\left(\mathcal{A}(\mathbf{h}_{i}^{L})\right). 
    \end{equation}

\subsection{Reinforcement Learning Control Policy}
\label{sec:rl}
    In this paper, the eventual goal is to learn a control policy for an in-hand manipulation task. Here, we formulate the control problems as an infinite-horizon discounted Partially Observable Markov decision process (POMDP). We define the state space or observation space as $S$ and the action space as $A$. To interact with the environments, the agent generates its stochastic policy $\pi_{\theta}(a\mid s)$ based on the current state $s$, where $a$ is the action and $\theta$ are the parameters of the policy function. The environment, on the other hand, produces a reward $r(s, a)$ for the agent, and the agent’s objective is to find a policy that maximizes the expected reward. Note that the state space $S$ is partially observed, i.e., only the robot hand-related states are observed, including robot joint states and tactile signals. 
    
    To maximize the expected reward, we use Proximal Policy Optimization (PPO) as our learning algorithm~\cite{fujimoto2018addressing}. We concatenate the predicted object states by perception model in Section~\ref{sec:TacGNN} and robot hand proprioceptive states (joint angle and velocity) as input. Output is the joint position control command. 
    
    In RL algorithm, RLU is used as an activation function. MLP architecture for action and value networks are [512, 256, 128]. We use Adam optimizer with a learning rate of $0.001$.

\subsection{Workflow}
\label{sec:workflow}

    Our framework consists of two stages - Perception and Controller. In the beginning, the perception model is initialized by Kaiming initialization~\cite{he2015delving} and frozen. Then, the control policy is trained with a Reinforcement Learning Algorithm. The input states consist of robot finger angles, robot finger velocity, object states inferred by the perception model, and the last action executed. Output is a position-controlled command. The reward function is:
    \begin{equation}
        R(s,a) = 0.5\times R_{\text{angle}} + \\
        250\times R_{\text{success}} + (-100)\times R_{\text{fail}} ,
    \end{equation}
    \noindent
    where $R_{\text{angle}} = \|\text{Angle}_{\text{now}} - \text{Angle}_{\text{pre}}\|$ denotes the angular change of two balls, $R_{\text{success}} = (\text{Angle}_{\text{now}}>180)$ denotes whether the task succeeds, $R_{\text{fail}}$ denotes whether the ball falls (out of the control of hand) or the time is out (the max time steps is 200).
    
    All the tactile information and object states are recorded in this stage. Once we get enough tactile data (25,000 steps), the perception model is trained using this dataset by supervised learning. Input states are the tactile information and output is the predicted object states. The loss function is computed by Root Mean Square Error (RMSE). 
    
    \begin{equation}
        RMSE = \sqrt{\frac{\sum_{t=1}^{T} (\hat{o_t} - o_t)^2 }{T}}, 
    \end{equation}
    where $T$ is the dimension of the object state, which is $6$ in this task. $\hat{o}$ is the predicted output and $o$ is corresponding state value.
    After 10 epochs of training, the method executes the reinforcement learning stage. These two stages are repeated till a maximum iteration number.

\section{Experiments}
\label{sec:experiment}

\subsection{Allegro hand with distributed sensors}
\label{sec:setup}
    An Allegro hand\footnote{https://www.wonikrobotics.com/research-robot-hand}, a four-fingered anthropomorphic robotic hand with four joints on each finger is used. Hence, the whole system has 16 DoFs.
    
    All the fingers and the palm have been covered with our in-house custom-made resistive tactile sensors~\cite{ding2021sim}. Fig.~\ref{fig:overview}(B) illustrates the distribution of the sensing elements (taxels). On each finger, the fingertip is covered by a 12$\times$6 taxel array, while inner-phalange links each have a 6$\times$ 6 taxel array. The palm is covered by 113 taxels. The readout from each taxel is one dimensional, corresponding to the normal force experienced by the taxel. The whole robot system setup is shown in Fig~\ref{fig:overview}. Two balls that have identical sizes are used as manipulated objects. 
    
    Simulation experiments are conducted in Isaac simulation~\cite{makoviychuk2021isaac}. We build simulation models based on the real system for the sim-to-real problem. All states in the simulation are added 10\% noise. The GNN model is built with Pytorch-Geometric~\cite{Fey/Lenssen/2019} and runs on Ubuntu 18.04 with a V100 GPU. All experiments are repeated three times with different random seeds.
    
\subsection{Object State Prediction}
    \label{sec:common methods}

    In this section, we focus on learning tactile feature and interpreting it into object states with supervised learning. The input is tactile sensing information and the output is the position and orientation of an in-hand object.
    
    \begin{table}[htbp]
        \centering
        \renewcommand\arraystretch{1.5}
        \caption{Experiment Results of Objects State Prediction.}
        \label{tab:state prediction}
        \resizebox{0.48\textwidth}{!}{
        
        \begin{threeparttable}
        \begin{tabular}{cccccc}
        \hline
        Object& Metric & MLP & CNN & GCN & TacGNN \\ 
        \hline
        \multirow{2}{*}{Cube} & Position Loss (cm) & 0.187 & 0.891 & 0.195 & \textbf{0.176}\\
         & Orientation Loss (°) & 12.31 & 15.87 & 10.33 & \textbf{9.17} \\
        \hline

        \hline
        \multirow{2}{*}{Cylinder} & Position Loss (cm) & 0.142 & 1.424 & 0.114 & \textbf{0.067}\\
         & Orientation Loss (°) & 14.25 & 27.76 & 12.32 & \textbf{8.76} \\
        \hline

        \hline
        \multirow{2}{*}{Meat Can} & Position Loss (cm) & \textbf{0.134} & 3.172 & 0.154 & 0.142\\
         & Orientation Loss (°) & 19.57 & 20.42 & 10.11 & \textbf{9.76} \\
        \hline

        \end{tabular}%
        
        \end{threeparttable}
        }
    \end{table}
    
    To validate the effectiveness and efficiency of our method for extracting tactile features, we conduct an object state prediction task on some regular shaped items - Cube, Cylinder, and the meat can from the YCB object set~\cite{calli2015ycb}. To get a contact-rich dataset, we execute a power grasp 200,000 times for each object with the random initial pose. Only the final step of each execution is recorded and the recording data includes tactile signals and object position and orientation. Note that in some cases, the pose of the object is rotational symmetrical and cannot be perceived by tactile sensing. For example, the haptic sensor will get the exact same signal when a square rotates 90 degrees around itself. So, we constrain the rotation angle of the cube to [0,90]. Similarly, the rotation angle of the cylindrical axis will be ignored.

    To prove the superiority of our method, we compare our proposed method to other commonly-used methods: \textbf{MLP} models directly take the 653 tactile values as input and predict the states of the target object; \textbf{CNN} models first reshape the tactile signals to a rectangular array and take it as input. \textbf{GCN} takes the tactile signals as a static graph. It first constructs the graph based on the initial pose of each sensor using the same method as TacGNN and the graph structure will not change over time.\textbf{TacGNN} only takes the activated tactile signal as input and the graph will be reconstructed at every step based on the interaction state as shown in Section~\ref{sec:method}. 
    
    For each experiment, 80\% dataset is used for training, and 20\% is used for the test. Root Mean Square Error (RMSE) is used to evaluate the accuracy of prediction. The results are shown in Table~\ref{tab:state prediction}. We can see that the proposed method achieves a better performance than other methods on all three datasets. MLP and CNN get worse results, especially on the cylinder and meat can.
    
    \begin{table}[htbp]
        \centering
        \renewcommand\arraystretch{1.5}
        \caption{Experiment Results of Balls Position Prediction.}
        \label{tab:Learning Tactile}
        \resizebox{0.48\textwidth}{!}{
        
        \begin{threeparttable}
        \begin{tabular}{cccc}
        \hline
        \textbf{Network}&\textbf{Architecture}&\textbf{Train Loss (cm)} &\textbf{Test Loss (cm)} \\ 
        \hline
        \textbf{MLP}& [64,64,64]&0.230&0.250\\ 
        \textbf{CNN+MLP}&[32,16],[784,128]&0.114&27.76\\ 
        \textbf{GCN+MLP}&[32,32,32],[32,6]&0.130&0.152\\ 
        \textbf{TacGNN+MLP}&[32,32,32],[32,6]&\textbf{0.077}&\textbf{0.096}\\ 
        \hline
        \textbf{TacGNN-32}&[32,32,32],[32,6]&0.077&0.096\\ 
        \textbf{TacGNN-16}&[16,16,16],[16,6]&0.340&0.251\\ 
        \textbf{TacGNN-64}&[64,64,64],[64,6]&0.098&0.085\\ 
        \textbf{TacGNN-128}&[128,128,128],[128,6]&\textbf{0.069}&\textbf{0.077}\\ 
        \hline
        \end{tabular}%
        
        \end{threeparttable}
        }
    \end{table}
    
\subsection{In-Hand Manipulation}
\label{sec:manipulation}
        
    \begin{figure*}
    \centering
    \includegraphics[width=0.9\textwidth]{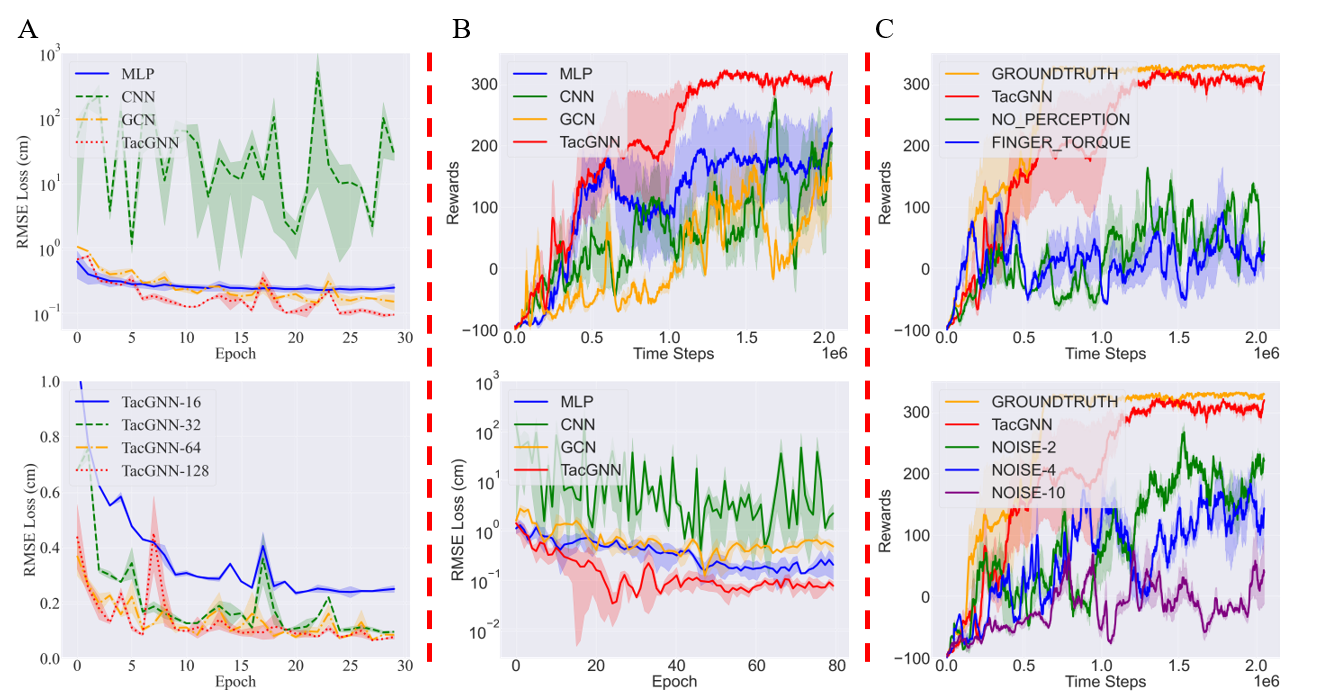}
    \caption{Learning Results of In-hand manipulation task. (A) Balls position prediction RMSE loss (cm) with different perception models - MLP, CNN, GCN, TacGNN (upper), and different layer sizes (bottom). (B) Baoding balls task learning results with MLP, CNN, GCN, TacGNN as perception model (upper) and their corresponding prediction RMSE loss in the process of training. (bottom) (C) Baoding balls task learning results with different inputs. Error margins indicate the standard deviation of three repeatable training.}
    \label{fig:TrainResults1}
    \vspace{-2em}
    \end{figure*}

    In this section, we consider an in-hand manipulation task- the Baoding balls task. Baoding balls task refers to the task of simultaneously manipulating two spheres to move around each other in the hand. Unlike the previous work~\cite{nagabandi2020deep}, we only take the robot hand states as input - robot joints states and tactile sensors signals. This task is challenging since the two balls share a compact workspace with each other which truly tests the dexterity of the manipulator. Also, the inter-object contact is difficult to be estimated and inferred.
    
    In this task, we first predict the position of two balls based on tactile sensing. Then, the predicted results and other robot states are taken as input to train a control policy. 
    
    \subsubsection{Perception Model with Different Architecture}

    To explicitly compare the perceived effects of different methods, the perception model is separated from the whole training process. We train the perception model with the same methods as before. The input of the model is tactile information and the output is balls object position $O = [P_1,P_2], P\in R^3$.

    To get a high-quality, i.e. rich tactile contact, dataset, we first train a policy to rotate two balls in hand simultaneously with precise ball positions. This also validates the feasibility of the task, as long as the object location is accurate enough. With this trained model, we get 20000 trajectories, each of which consists of around time-indexed 1000000 steps. In each step, raw tactile data is recorded as input and the position of two balls is recorded as the output label. 

    The results are shown in Fig.~\ref{fig:TrainResults1} (A) and Table~\ref{tab:Learning Tactile}. We could see that our method TacGNN performs the lowest Root Mean Square Error (RMSE). In addition, CNN has a much higher RMSE on the test dataset than the training dataset, which means CNN is more likely to overfit after training.

    To get an optimized network architecture for the downstream manipulation task, we change the channels, i.e. the layer size of the network, to 16, 64, 128 as shown in Fig.~\ref{fig:TrainResults1} (A) and Table~\ref{tab:Learning Tactile}. The results show that 32 channels are enough considering the prediction accuracy and the consumption of computational resources.
    
    \begin{table}[htbp]
        \centering
        \renewcommand\arraystretch{1.5}
        \caption{Success rate of different perception models on in-hand manipulation task.}
        \label{tab:TrainResults1}
        \resizebox{0.48\textwidth}{!}{

        \begin{tabular}{cccccc}
        \hline
        &\textbf{Task}&\textbf{MLP}&\textbf{CNN}&\textbf{GCN}&\textbf{TacGNN} \\
        \hline
        &\textbf{Simple}& $52.44$ & $42.71$ & $70.62$ & $\textbf{94.71}$ \\

        \cline{2-6}
        \textbf{Success Rate}&\textbf{Middle}& $45.57 $ & $63.75$ & $64.51$ & $\textbf{88.57}$\\ 
        
        \cline{2-6}
        &\textbf{Hard}& $41.78 $ & $50.88$ & $55.11$ & $\textbf{79.78}$\\ 

        \hline
        &\textbf{Simple}& $\textbf{76.41} $ & $146.71$ & $99.79$ & $82.14$\\ 
        
        \cline{2-6}
        \textbf{Average Steps}&\textbf{Middle}& $\textbf{77.47} 2$ & $132.51$ & $86.83$ & $86.11$\\ 
        
        \cline{2-6}
        &\textbf{Hard}& $\textbf{79.03}$ & $119.11$ & $102.79$ & $93.93$\\ 
        \hline
        \end{tabular}%
        }  
    \end{table}

    \subsubsection{Learning Control Policy with Different Perception Model}
    \begin{figure*}[thpb]
      \centering
      \includegraphics[width=1.0\textwidth]{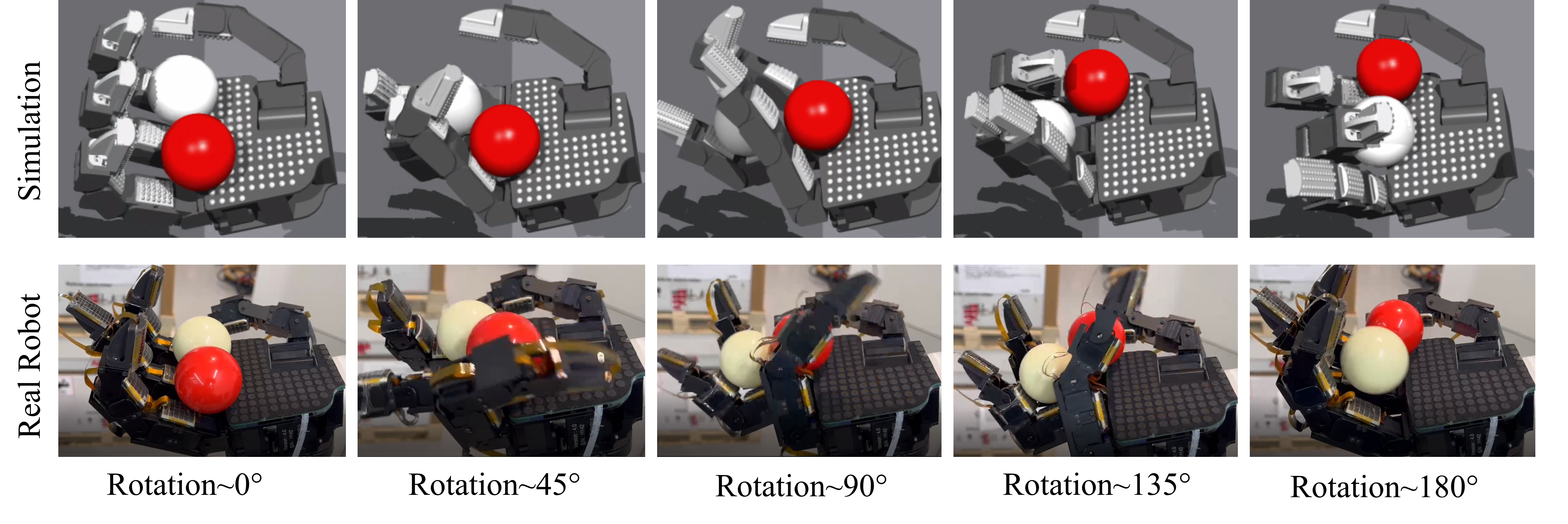}
      \caption{Key frames of in-hand manipulation experiments on simulation and real robot.}
      \label{fig:Keyframes}
      \vspace{-2mm}
   \end{figure*}
   
    In this section, we train a control policy with different perception models: MLP, CNN, GCN and TacGNN. The perception model architecture for all methods is the same as before.
    
    As shown in Fig.~\ref{fig:TrainResults1} (B), the proposed method outperforms other methods in the whole process of training. Other methods can achieve good performance in some cases, but with large variance. Fig.~\ref{fig:TrainResults1} (B)show the prediction loss (RMSE Loss) during the training. Our proposed method always gets the lowest loss as expected. The In-hand manipulation task reward positively correlated with the performance of the perception model.
    
    We further test these trained models on the tasks with three difficulty levels without further training.

    \begin{enumerate}
    \item Simple task - rotate two balls simultaneously around each other at more than 180 degrees in angle which is the original task in the training process.
    
    \item Middle task - rotate two balls simultaneously around each other at more than 180 degrees in angle with a disturbing force every ten steps. The magnitude of force $ F\sim\mathcal N(0,0.1)$ N and the direction is totally random.
    
    \item Hard task - In addition to the middle task, this task requires that the robot hold for 5 steps after completing the rotation task, in case the robot accidentally achieves the rotation goal. This task is challenging since the robot has to keep the ball from falling after the completion of the task.
    \end{enumerate}

    As shown in Table~\ref{tab:TrainResults1}, TacGNN gets the highest success rate in all three tasks. In particular, our method achieves around an 80 percent success rate in hard tasks. Note that the middle task and hard task is a unseen scenario for trained policy and no further training needs to be done. As for the average steps cost in each trial, the MLP method always took the least amount of steps, but the lower success rate suggests that the falling of balls causes the experiment to end early.
    
    \subsubsection{Learning Control Policy with Different Input States}

    To investigate the influence of tactile information on the final performance of in-hand manipulation, we conduct experiments by changing the input state types. \textbf{GROUNDTRUTH}: The input includes precise object positions and robot joint angles and velocity. \textbf{NO\_PERCEPTION}: The input only includes robot joint angles and velocity. \textbf{FINGER\_TURQUE}: The input includes robot joint angles, velocity and torques. \textbf{NOISE\_2,4,10}: The input includes object positions with a uniform random noise [-2,2], [-5,5], [-10,10] mm to simulate perception error due to obscuration, calibration error, etc., and robot joint angles and velocity.

    As shown in Fig.\ref{fig:TrainResults1} (C), TacGNN could eventually catch up to Groundtruth's performance with more training data. NO\_PERCEPTION and FINGER\_TURQUE remain the low rewards since it lacks enough interaction-related information. Noised object information influences the speed of convergence and final performance.

\subsection{Real Robot Experiment}

    Finally, we validate our method on a real robot system as shown in Fig.\ref{fig:overview}(B). Domain Randomization~\cite{tobin2017domain} is adapted to enable policy trained solely in simulation to transfer to a real robot. Domain randomisation parameters include Mass of ball~[0.1, 0.2]kg; Friction~[0.8,1.2]; Gravity [7.0,12.0]$m/s^2$; Joint Stiffness and Damping [0.8,1.2]$\times$ Initial Value. The initial value of joint stiffness and damping is determined by a grid search. First, the real robot executes a series of position-controlled commands ($0^{\circ},15^{\circ},30^{\circ},45^{\circ},60^{\circ}$) and the angles of joint are recorded. Then, a grid search - joint stiffness [1,3,5]$N/m$, joint damping [0,0.1,0.5]$Ns/m$ is conducted in simulation and the same position-controlled commands are executed for each joint. Finally, the recorded angles in simulation and reality are computed to a loss. The parameters with the lowest loss are chosen as the initial value of joint stiffness and damping. 

    A successful trial in the real robot is shown in Fig.\ref{fig:Keyframes}, the robot shows a similar performance to the simulation. This verifies that the learnt tactile feature and policy can be applied to a real robot. 

\section{Discussion}
\label{sec:discussion}

\subsubsection{Object States Perception}
    Robotic tactile sensing brings 3D physical relations between robots and objects. Compared to vectors and 2D images, graph data structure could  maintain the geometrical relations among data points. That could be the reason that GCN and TacGNN perform better in object states prediction tasks, especially for cylinder and meat can. 
    
    CNN reshapes the tactile array to a rectangular shape, which highly destroys the relations of tactile signals. For example, the tips of the thumb and middle finger will be close together when power grasp, but they will be far apart in the reshaped picture. This leads to the worst performance and instability of CNNs in prediction tasks.
    Compared to GCN, TacGNN only takes the activated tactile nodes as input and reconstructs the graph at every step, which is beneficial for encoding the spatial relationship from diverse hand poses.
    
\subsubsection{In-hand Manipulation}

    As for the comparison experiments on perception methods, the reward positively correlated with the performance of the perception model. The control policy with TacGNN gets the best rewards and the fastest convergence time.

    More than the basic test, simply rotating two balls more than 180 degrees, we add two advanced tasks with external perturbations and hold-on targets. Our method handles well and is generalized to these without further training. 

    As for the comparison experiments on the input information, we can see that although GROUNDTRUTH converge faster than TacGNN, our method can achieve similar performance with enough training data. Object-hand interaction information is essential for this in-hand manipulation task. To a certain extent, we accomplished this dexterous manipulation task by relying on the robot hand's own perception.
    
\subsubsection{Small Sim2Real Gap}
    
    We get similar performances in simulation and reality as shown in Fig.~\ref{fig:Keyframes}. The first frame is the initial pose of the robot and object. Then, the middle finger stretches, and the ring finger compresses to stop the red ball from falling. In the next few frames, the index finger, middle finger, and ring finger construct a concavity to contain the white ball. Finally, the red ball is pushed to the other side by the ring finger and the white ball is pushed to the middle of the palm. The thumb finger is another "wall" to stop balls from falling.

\section{Conclusion and Future Works}
\label{sec:conclusion}

    In conclusion, this paper presents a novel framework for tactile-based in-hand manipulation tasks. First, we propose a novel method-TacGNN as a perception model to extract tactile features-manipulated object states. Then, using the tactile features from the perception model, we train an autonomous control policy with PPO reinforcement learning algorithm for an in-hand manipulation task. The results show that our method outperforms other commonly-used methods (MLP, CNN, GCN) to interpret tactile information. Finally, features extracted with our perception model are able to provide enough interaction information for in-hand manipulation tasks and such dexterous skills could be achieved solely relying on robotics own perception. We validate our approach by implementing it on a real robot system and achieve comparable performance results.
    
    However, this work is a preliminary exploration of a GNN-based tactile in-hand manipulation task. There are limitations in this work. The ”tactile blind area" occurs in manipulation. For example, the object could be stuck in the gap between tactile arrays. At that step, no tactile sensors are activated and no tactile sensing is input. A possible solution is to use time-series information so that robot could deduce an approximate perception of objects from previous data.

\bibliographystyle{IEEEtran}
\bibliography{reference}

\end{document}